\documentclass[10pt,twocolumn,letterpaper]{article}

\usepackage{cvpr} % uncomment this for the final submission
\usepackage{times}
\usepackage{epsfig}
\usepackage{graphicx}
\usepackage{amsmath}
\usepackage{amssymb}
\usepackage{multirow}
\usepackage[numbers]{natbib}

\begin{document}

%%%%%%%%% TITLE
\title{M3DHMR: Monocular 3D Hand Mesh Recovery}

\author{Yihong Lin\\
South China University of Technology\\Guangzhou, China\\
{\tt\small jlxxlyh@163.com}
\and
Xianjia Wu\\
Huawei Cloud\\
Shenzhen, China\\
{\tt\small wuxianjia1996@gmail.com}
\and
Xilai Wang\\
South China University of Technology\\Guangzhou, China\\
{\tt\small auwangxilai@mail.scut.edu.cn}
\and 
Jianqiao Hu\\
South China University of Technology\\Guangzhou, China\\
{\tt\small 
jianqiaohu2000@163.com
}
\and
Songju Lei\\
Nanjing University\\Nanjing, China\\
{\tt\small
leisongju@smail.nju.edu.cn
}
\and 
Xiandong Li$^{\dagger}$\\
Huawei Cloud\\
Shenzhen, China\\
{\tt\small
lxdphys@smail.nju.edu.cn
}
\and
Wenxiong Kang$^{\dagger}$\\
South China University of Technology\\Guangzhou, China\\
{\tt\small 
auwxkang@scut.edu.cn
}
}

\maketitle
\thispagestyle{empty}

%%%%%%%%% ABSTRACT
\begin{abstract}
   Monocular 3D hand mesh recovery is challenging due to high degrees of freedom of hands, 2D-to-3D ambiguity and self-occlusion. Most existing methods are either inefficient or less straightforward for predicting the position of 3D mesh vertices. Thus, we propose a new pipeline called Monocular 3D Hand Mesh Recovery (M3DHMR) to directly estimate the positions of hand mesh vertices. M3DHMR provides 2D cues for 3D tasks from a single image and uses a new spiral decoder consist of several Dynamic Spiral Convolution (DSC) Layers and a Region of Interest (ROI) Layer. On the one hand, DSC Layers adaptively adjust the weights based on the vertex positions and extract the vertex features in both spatial and channel dimensions. On the other hand, ROI Layer utilizes the physical information and refines mesh vertices in each predefined hand region separately. Extensive experiments on popular dataset FreiHAND demonstrate that M3DHMR significantly outperforms state-of-the-art real-time methods. The code is available at https://github.com/Jackson-coder/M3DHMR.
\end{abstract}

%%%%%%%%% BODY TEXT
\section{Introduction}

Monocular 3D hand mesh recovery has attracted great attention in VR/AR, human-computer interaction, etc \cite{tang2021towards, 2020MEgATrack}. This task aims to estimate the 3D spatial positions of hand mesh vertices from a single RGB image. It improves the accuracy and security of hand gesture authentication by restoring biometric features. On the one hand, it repairs low-quality samples caused by factors such as wear, contamination, or limitations of capture devices, which reduces false recognition rates. On the other hand, it reconstructs incomplete data, thereby boosting performance of multimodal hand gesture authentication system. However, high degrees of hand freedom, 2D to 3D ambiguity and self-occlusion make significant challenges for hand reconstruction \cite{boukhayma20193d, ge20193d, Chen_Liu_Ma_Chang_Wang_Chen_Guo_Wan_Zheng_2021}.

To overcome these challenges, parametric model-based approaches \cite{zhou2020monocular,chen2021model,seeber2021realistichands,pavlakos2024reconstructing} usually use MANO \cite{Romero_Tzionas_Black_2017} which obtains a hand mesh from coefficients of shape and pose to facilitately generate diverse hand poses for various applications with flexibility and high-precision. Though these methods are flexible, their 3D structure is not intuitive enough, especially the shape parameters are too difficult to optimize. Conversely, vertex-based methods can directly estimate the coordinates of the 3D mesh. Specifically, these models usually define the wrist of the hand as the root and calculate the root-relative 3D mesh coordinates. Some latest researches \cite{lin2021end, lin2021-mesh-graphormer, PointHMR} are unable to implement in real-time due to their exclusive focus on improving the reconstruction accuracy, without considering the scale of models. Others \cite{MobileHand:2020, kulon2020weakly, Choi_2020_ECCV_Pose2Mesh, zhang2021hand, chen2021i2uv, tang2021towards,guo_clip_hand,vasufastvit2023} can achieve real-time performance but are limited in reconstruction accuracy. Additionally, all methods previously mentioned have to provide the absolute root coordinate in camera space to get absolute camera coordinates of the 3D mesh, which limits the application to many high-level tasks. Though CMR \cite{Chen_Liu_Ma_Chang_Wang_Chen_Guo_Wan_Zheng_2021} and MobRecon \cite{Chen_Liu_Dong_Zhang_Ma_Xiong_Zhang_Guo_2022} provide pipelines to directly acquire the absolute 3D coordinates of the mesh in real-time, their reconstruction accuracy still needs to be improved.

\begin{figure}[t] %!t
\centering
\includegraphics[scale=0.53]{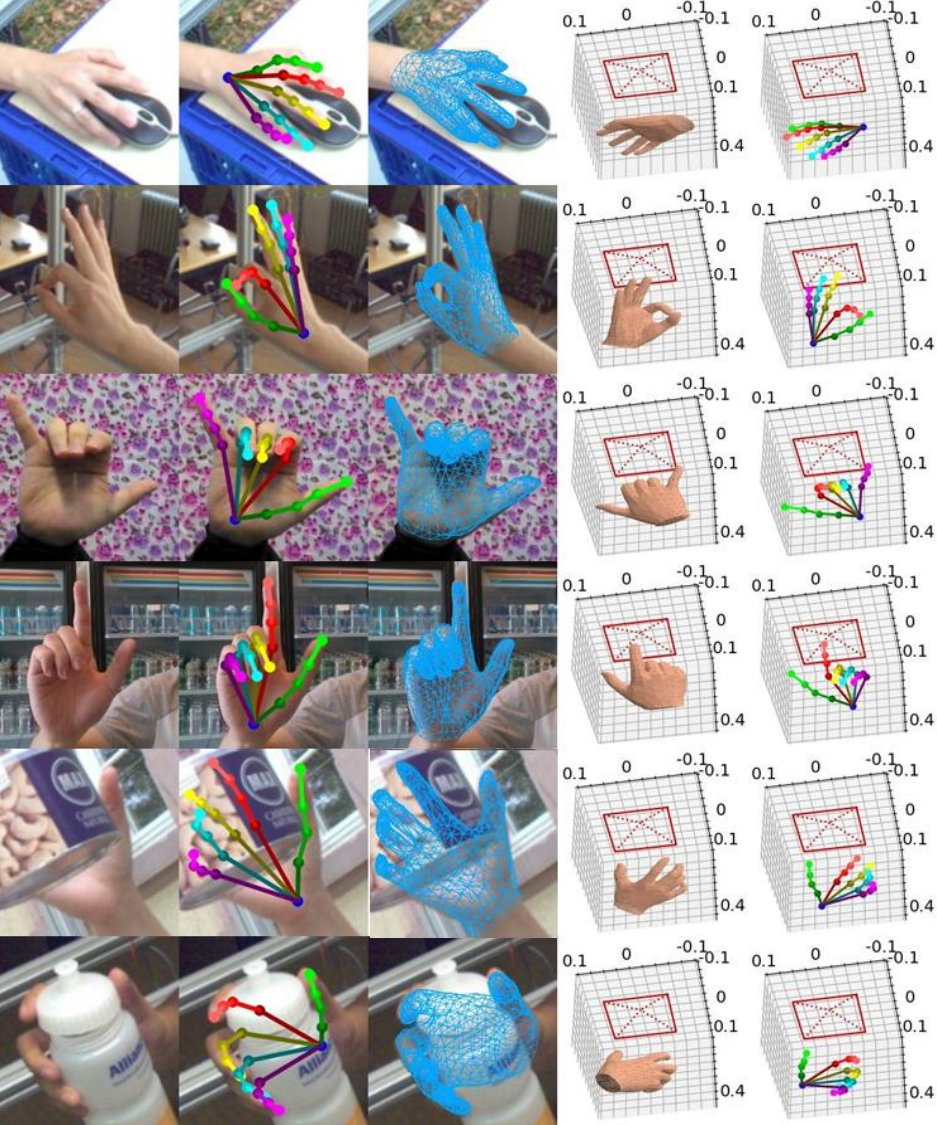}
\caption{Qualitative results of our M3DHMR. We show the 2D pose, mesh projection, camera space mesh and pose in metres. The red rectangle indicates the camera.}
\label{fig1}
% \vspace{-1.0em}
\end{figure}

Herein, we propose a Monocular 3D Hand Mesh Recovery (M3DHMR) pipeline to further improve the performance of real-time camera-space 3D hand mesh recovery. The pipeline reconstructs the 3D hand through multiple stages and facilitates the 3D reconstruction by regressing 2D joint landmarks as a subtask. Novel Dynamic Spiral Convolution (DSC) Layers and Region of Interest (ROI) Layer are used to decode the 3D vertex features to obtain mesh 3D coordinates. Instead of employing stacked spiral convolution layers or increasing the number of channels in these layers to enhance the model capacity, which is often accompanied with expensive computation, DSC Layers explore a more reasonable spiral convolution architecture to further enhance the model's representation capability with minimal increase in number of parameters and computational cost. More concretely, two convolution layers of DSC are used to adaptively adjust the normalized weights of the $K$ spiral convolutions to extract the spatial-channel information. Furthermore, ROI Layer introduces the spatial information of the hand in the physical world by dividing the hand into multiple physical regions and refining the hand mesh vertices in each region separately. Extensive experiments indicate that M3DHMR surpasses current state-of-the-art real-time models in 3D hand mesh reconstruction. Some reconstruction results of M3DHMR in situations of complex pose, self-occlusion and object occlusion are shown in Fig. \ref{fig1}. Our contributions are summarized as follows:

\begin{itemize}
\item We design DSC Layers to adapt the weights of spiral convolutions based on the vertex positions and explore the mesh features in both space and channels.
\item We propose a ROI Layer that divides the mesh vertices of the hand into several sets based on the physical region to separately refine the hand mesh features.
\item Extensive experiments demonstrate that our method achieves superior performance in terms of reconstruction accuracy over existing state-of-the-art real-time methods.
\end{itemize}

\section{Related Work}
\subsection{Hand Mesh Estimation}
Hand mesh estimation has become an important task in computer vision and augmented reality, which aims to reconstruct detailed 3D hand structures from monocular RGB images. Existing mainstream methods can be categorized into model-based and vertex-based approaches.

Model-based approaches commonly rely on parametric hand models, such as MANO \cite{Romero_Tzionas_Black_2017}, which provides a low-dimensional representation of hand poses and shapes. Building upon this foundation, Zhou et al. \cite{zhou2020monocular} proposed the first real-time monocular hand capture framework, integrating multi-modal data (RGB, keypoints, and contours) to achieve highly accurate 3D reconstructions. Furthermore, Chen et al. \cite{chen2021model} developed a self-supervised model-based 3D hand reconstruction method that eliminates the need for 3D annotation by leveraging geometric constraints and 2D-3D consistency for training. Meanwhile, RealisticHands \cite{seeber2021realistichands} combined parametric mesh priors with non-parametric details, enabling high-quality hand pose and shape estimation from monocular RGB inputs. HaMeR \cite{pavlakos2024reconstructing} used a fully transformer-based architecture to analyze hands with significantly increased accuracy and robustness. Despite their efficiency, these methods are inherently limited by the expressiveness of parametric models.

Vertex-based methods directly predict the 3D coordinates of mesh vertices, offering flexibility beyond parametric models. Among these, Mesh Graphormer \cite{lin2021-mesh-graphormer} and PointHMR \cite{PointHMR} pioneered the use of transformer architectures for monocular 3D hand mesh recovery, achieving notable improvements in accuracy and robustness. However, their computational complexity hinders real-time deployment. To address this limitation, several efficient alternatives have emerged. MobileHand \cite{MobileHand:2020} introduced a lightweight network design, while YoutubeHand \cite{kulon2020weakly} and Pose2Mesh \cite{Choi_2020_ECCV_Pose2Mesh} leveraged graph convolutions for efficient mesh reconstruction. Further explorations include HIU-DMTL \cite{zhang2021hand}, which employed deep multi-task learning, and I2UV-HandNet \cite{chen2021i2uv}, which adopted an image-to-UV mapping strategy. Tang et al. \cite{tang2021towards} enhanced real-time performance and significantly improved pose and shape accuracy through a novel alignment optimization module. CMR \cite{Chen_Liu_Ma_Chang_Wang_Chen_Guo_Wan_Zheng_2021} and MobRecon \cite{Chen_Liu_Dong_Zhang_Ma_Xiong_Zhang_Guo_2022} proposed end-to-end pipelines to directly estimate absolute 3D coordinates. More recently, CLIP-Hand3D \cite{guo_clip_hand} integrated vision-language pretraining with context-aware prompting to improve generalization in challenging scenarios, and FastViT \cite{vasufastvit2023} introduced a hybrid vision transformer with structural reparameterization for optimal speed-accuracy trade-offs.

\begin{figure*}[t] %!t
\centering
\includegraphics[scale=0.74]{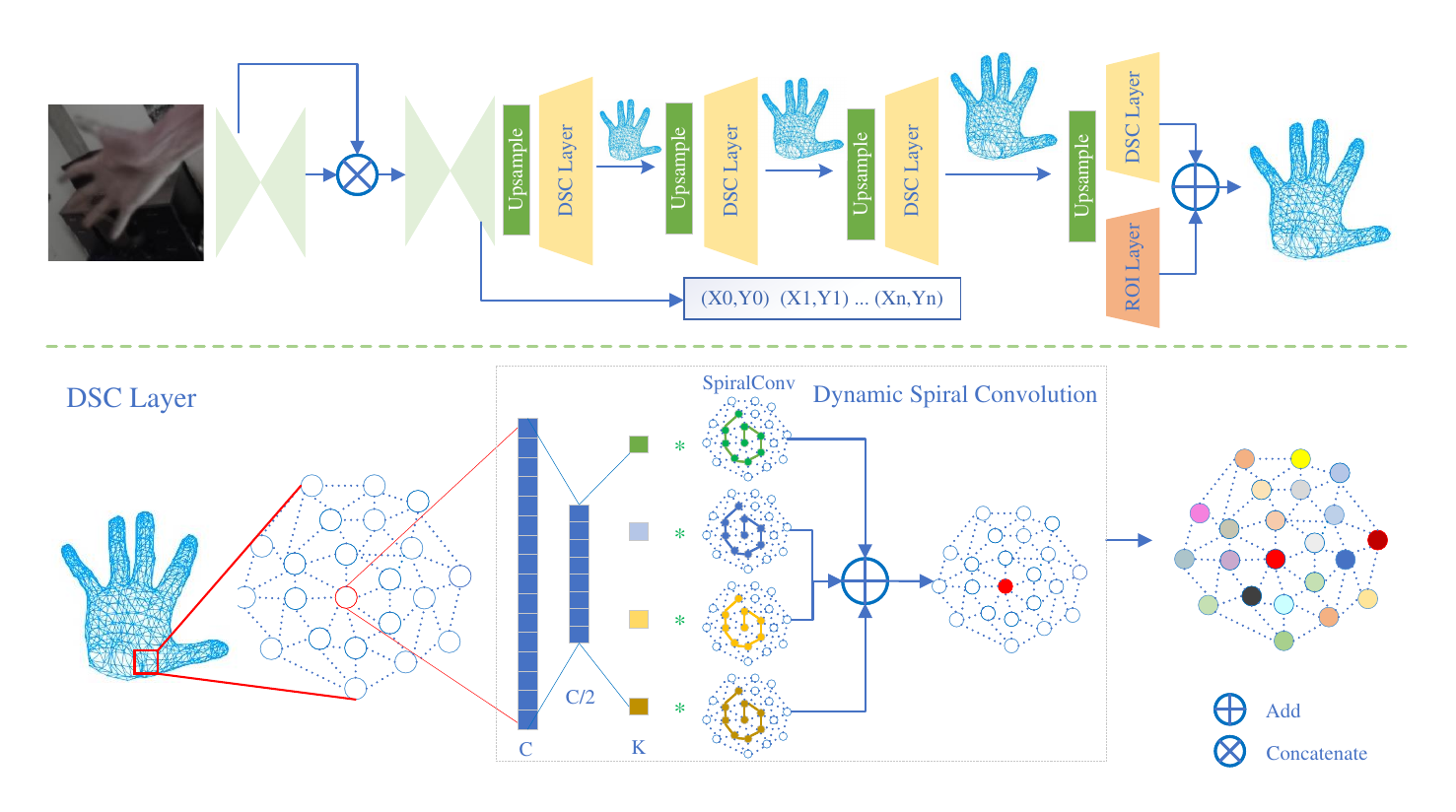} %0.63
\caption{Overview of our M3DHMR framework. The architecture of ROI Layer is a dilated spiral convolution. Different colors in Dynamic Spiral Convolution indicate different weights corresponding to the spiral convolution kernels.}
\label{fig2}
% \vspace{-1.0em}
\end{figure*}

\subsection{Spiral Convolution}

Defferrard et al. \cite{defferrard2016convolutional} pioneered graph convolutional networks (GCNs) that operate on non-Euclidean graph structures while maintaining computational complexity comparable to that of classical CNNs. Masci et al. \cite{masci2015geodesic} developed spatial graph convolution operators that incorporate geodesic-aware sampling of graph signals, providing more intuitive handling of 3D mesh vertices in the spatial domain. This line of research was further advanced by Lim et al. \cite{lim2018simple}, who introduced spiral convolution based on graph convolution as an effective approach for processing 3D mesh vertices.

Subsequent improvements in mesh spatial convolution techniques include SpiralNet++ \cite{gong2019spiralnet++}, which employed truncated spiral lines to limit sampled vertices while expanding the receptive field through hollow spiral lines. Kolotouros et al. \cite{kulon2020weakly} directly learned the mapping relationships from 2D images to 3D meshes by implementing a straightforward encoder-decoder architecture with spiral convolution. Recent works like CMR and MobRecon designed ISM and DSConv to further enhance feature extraction.

Despite these advancements, the weights of current methods remain fixed when training is finished. This fixed nature restricts their ability to adapt to varying spatial locations of mesh vertices. To solve this problem, our work explores a more flexible convolution operator that can dynamically adjust to the spatial distribution of mesh vertices, potentially offering improved performance in 3D mesh processing tasks.

\section{Method}
\subsection{Overview}
Taking a single RGB image as input, our goal is to get the 3D coordinates of the hand mesh vertices. M3DHMR firstly uses a two-stack hourglass network \cite{newell2016stacked} to generate encoding features $F^e$ and predict 2D joint landmarks $L^p$. Note that the first stack and the first layer of the second stack are pretrained under guidance of heatmaps and landmarks. Moreover, M3DHMR adopts a spiral decoder to infer the root-relative coordinates of the 3D mesh vertices from the encoding features $F^e$. Specifically, pose features $F^p$ are first obtained by grid sampling:
\begin{align}
    F^p = F^e(L^p).
\end{align}%
Then a learnable matrix $M$ lifts 2D pose features to 3D mesh features: 
\begin{align}
    F^m = M \cdot F^p.
\end{align}%
Subsequently, multiple graph-based DSC Layers produce multiple scale mesh vertex maps in a coarse-to-fine manner. Each layer upsamples output mesh features from the previous layer and explores spatial connectivity and channel dependencies between vertices to get fine-grained vertex features. Finally, M3DHMR adopts a ROI Layer to refine the result and uses the same method as CMR \cite{Chen_Liu_Ma_Chang_Wang_Chen_Guo_Wan_Zheng_2021} to estimate the position of root in the camera space and thus recovers the coordinates of the 3D hand mesh vertices in the camera space. The overall network structure is shown in Fig. \ref{fig2}. 

\begin{figure}[t]  %!t
\centering
\includegraphics[scale=1.15]{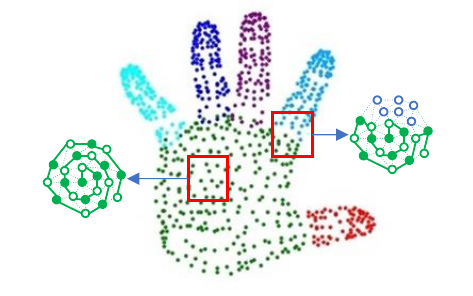}
\caption{The preset regions for the ROI Layer. Different colors denote different regions. Dilated
spiral convolution in ROI layer focuses only on the interior of the region.}
\label{fig3}
% \vspace{-1.0em}
\end{figure}

\subsection{Dynamic Spiral Convolution Layer}\label{AA}
Similar to SpiralConv \cite{lim2018simple} and SpiralConv++ \cite{Gong_Chen_Bronstein_Zafeiriou_2019}, dynamic spiral convolution determines the convolution centre and produces a sequence of enumerated centre vertices based on adjacency, followed by the 1-ring vertices, the 2-ring vertices, and so on, until all vertices containing n rings are included:
\begin{align} 
    0{-ring}(v) &= v, \nonumber \\
    (n+1){-}ring(v) &= \mathcal{N}(n{-ring}(v)) \setminus n{-disk}(v), \nonumber \\
    n{-disk}(v) &= \cup_{i=0,...,n}i{-ring}(v), 
\end{align}%
where $\mathcal{N}(V)$ selects all vertices in the neighborhood of any vertex in set $V$. Different from previous works, dynamic spiral convolution comprises $K$ convolution kernels that share the same kernel size and input/output dimensions. To enable adaptive adjustment of convolution weights for vertices at different positions, it employs two convolution layers and softmax to generate normalized attention weights for the $K$ convolution kernels:
\begin{align} 
    \pi_i(x) &= f_2(f_1(x)), \nonumber \\
    s.t. 0\leq \pi_i(x)& \leq1, \sum_{i=1}^K \pi_i(x)=1,
\end{align}%
where $f_1$ and $f_2$ are the two convolution layers, and $\pi_i$ is the attention weight for the $i$-th convolutional kernel. The initial convolution layer reduces the dimension to half, while the latter computes the normalized attention weights for the $K$ convolutional kernels, ensuring computational efficiency. Once the $K$ attention weights are obtained, the final convolution weight parameters for that position can be calculated by weighted summation:
\begin{align} 
    \hat{W}_1=\sum_{i=1}^K W_1, \hat{b}_1=\sum_{i=1}^K b_1, \nonumber \\
    \hat{W}_2=\sum_{i=1}^K W_2, \hat{b}_2=\sum_{i=1}^K b_2. 
\end{align}%
The final output can be denoted as:
\begin{align} 
    DSC(v)=\hat{W}_2(\hat{W}_1(f(k{-disk}(v)))+\hat{b}_1)+\hat{b}_2.
\end{align}%

\subsection{Region of Interest Layer}
The hand is artificially divided into six regions: palm, thumb, index finger, middle finger, ring finger, and little finger based on the shape and motion characteristics (See Fig. \ref{fig3}). We believe that the mesh vertices in the same region are highly correlated. The movement of the vertices strongly affects others within the same region but has less effect on vertices in other regions. For example, movement at the joint of the index finger causes significant displacement at the tip of the finger, but has minimal impact on the palm and other fingers. Therefore, we assume that for each vertex, the finger or palm region where it is located will be the region of interest. Benefiting from the large receptive field of dilated spiral convolution \cite{Gong_Chen_Bronstein_Zafeiriou_2019}, ROI Layer can sufficiently explore the global features in each hand region of convolution centre to refine the mesh in predefined hand regions.

\subsection{Loss Function}
We adopt $L_1$ loss to the 3D mesh loss $L_{mesh}$ and 2D pose loss $L_{pose2D}$. Inspired by \cite{ge20193d}, we adopt the normal loss $L_{norm}$ and the edge-length loss $L_{edge}$ to preserve the surface normals and penalize flying vertices. Additionally, we
conduct consistency supervision in both 3D and 2D space similar to \cite{guo2018stacked} and \cite{yang2021semihand}. Formally, we have
\begin{align} 
    \mathcal{L}_{mesh}&=||V-V^*||_1, \nonumber \\\mathcal{L}_{pose2D}&= ||L^p-L^{p,*}||_1,\nonumber \\ \mathcal{L}_{norm}&=\sum_{c\in C}\sum_{(i,j)\subset c}|\frac{V_i-V_j}{||V_i-V_j||_2}\cdot n_c^\star| \nonumber, \\
    {\mathcal L}_{edge}&=\sum_{c\in C}\sum_{(i,j)\subset c}|||V_{i}-V_{j}||_{2}-||V_{i}^{\star}-V_{j}^{\star}||_{2}|, \nonumber\\
    \mathcal{L}_{con3D}&=||V_{view1}-V_{view2}||_1, \nonumber\\
    \mathcal{L}_{con2D}&=||L^{p}_{view1}-L^{p}_{view2}||_1,
\end{align}%
where $C$, $V$ are face and vertex sets of a mesh; $n_c^\star$ indicates unit normal vector of face $c$; and $*$ denotes the ground truth.

Our overall loss is $\mathcal{L}_{total}=\mathcal{L}_{mesh}+\mathcal{L}_{pose2D}+\mathcal{L}_{norm}+\mathcal{L}_{edge}+\mathcal{L}_{con3D}+\mathcal{L}_{con2D}$.

\section{Experiments}

\subsection{Datasets and Metric}

\textbf{FreiHAND} is the current largest hand dataset for hand pose and shape estimation from single color images. It contains 130,240 training images and 3,960 evaluation samples. The data of FreiHAND is obtained from real-world scenarios in situations of complex pose, self-occlusion and object occlusion. Meanwhile, it is the only dataset that is annotated with both high-precision 3D keypoints and mesh models.
% \textbf{HO3Dv2} is a 3D hand-object dataset that contains 66,034 training samples and 11,524 evaluation samples. We use this dataset only for evaluation to demonstrate the superiority of our method.

We use the following metrics in quantitative evaluations.

\textbf{PA-MPJPE/MPVPE} ignores global variations and measures the mean per joint/vertex position error in terms of Euclidean distance (mm) between the root-relative prediction and the ground truth coordinates based on Procrustes analysis \cite{gower1975generalized}. For brevity, the metric is abbreviated as PJ/PV.

\textbf{F-Score} is the harmonic mean between precision and recall between two meshes \emph{w.r.t.} a specific distance threshold, which enables a balanced assessment of the model's performance, particularly useful when the distribution is unbalanced. F@5/F@15 denotes to a threshold of 5mm/15mm.

\textbf{FPS} measures the number of frames an algorithm can process per second. 
%Higher FPS means faster processing and lower latency.

\subsection{Experiment Setup}

We set the batch size to 32 and train 38 epochs with Adam optimizer \cite{kingma2014adam}. The learning rate is initialized to 0.001, which is divided by 10 at 30th epoch. Data augmentation includes random box scaling, box rotation, box shift and color jitter.
We use PyTorch as the framework to conduct all experiments and use a single NVIDIA RTX 3090 for training and a single NVIDIA RTX 2080 Ti for inference. 

\begin{table}[t]
\begin{center}
\begin{tabular}{|c|c|c|c|c|}
\hline
& PJ$\downarrow$ & PV$\downarrow$ & F@5$\uparrow$ & F@15 $\uparrow$  \\
\cline{1-5}
Baseline& 6.8& 7.0& 0.720&0.967 \\
\hline
\begin{tabular}[c]{@{}c@{}}w/o DSC Layer\end{tabular} &6.8& 6.9& 0.723&0.977 \\
\hline
w/o ROI Layer  &6.7& 6.8& 0.728& 0.978 \\
\hline
Ours  &\textbf{6.6}&\textbf{6.7}& \textbf{0.734}& \textbf{0.978} \\
\hline
\end{tabular}

\end{center}
\caption{Ablation study of DSC Layer and ROI Layer with Kernel Size of 8. "Baseline" denotes the model with only Spiral Convolution Layers.}
\label{tab1}
\end{table}

\begin{figure*}[t] %!t
\centering
\includegraphics[scale=0.55]{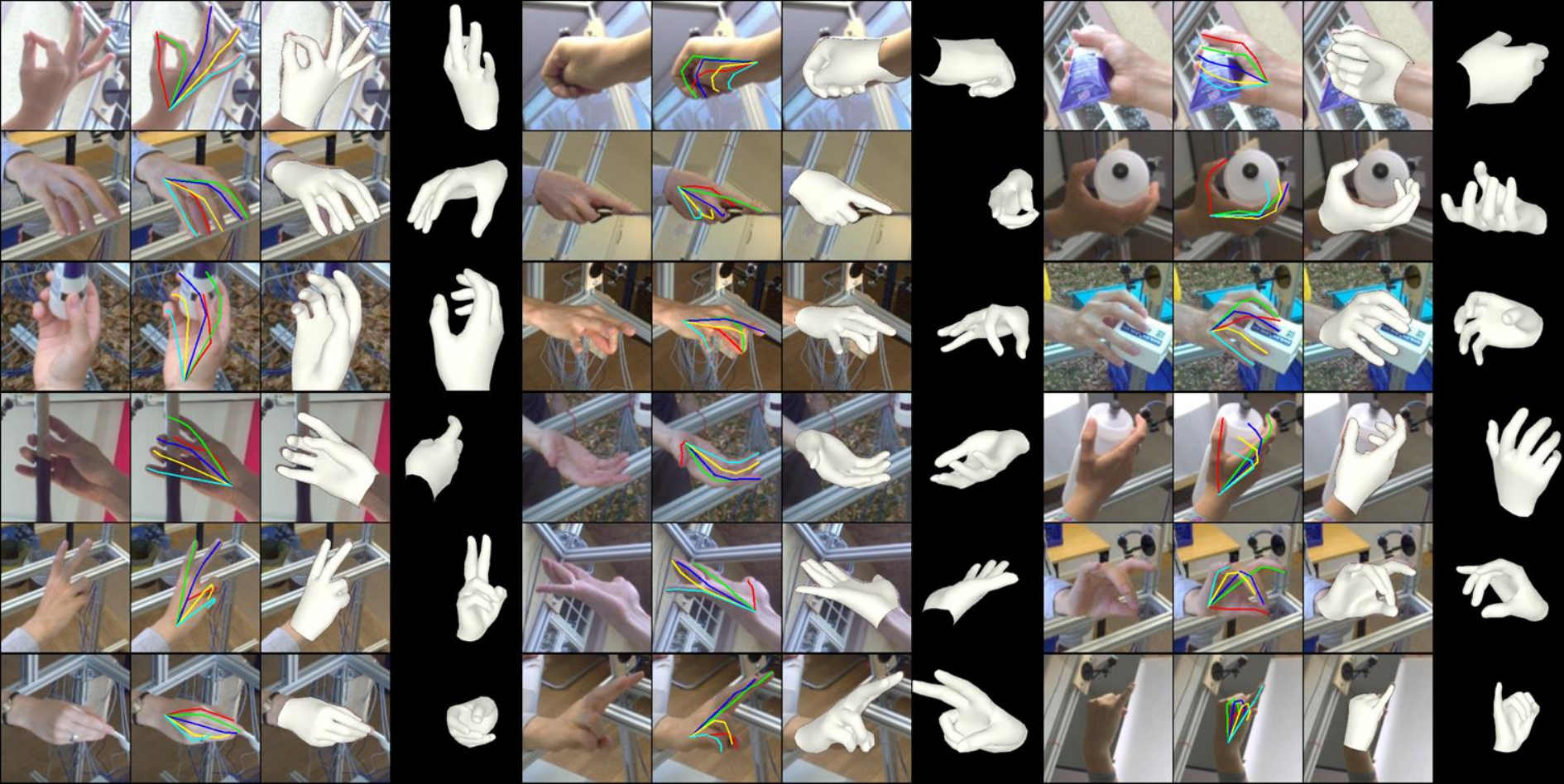}
\caption{Several qualitative results of our predicted 2D pose, front view mesh and side view mesh on the test set of FreiHAND.}
\label{fig4}
% \vspace{-1.0em}
\end{figure*}

\begin{table*}[t]
\begin{center}
\begin{tabular}{|c|c|c|c|c|}
\hline
3D decoding & PJ$\downarrow$ & PV$\downarrow$ & F@5$\uparrow$ & F@15 $\uparrow$  \\
\cline{1-5}
SpiralConv++\cite{Chen_Liu_Ma_Chang_Wang_Chen_Guo_Wan_Zheng_2021}& 6.8& 6.9& 0.723& 0.976\\
\hline
Depth-Separable SpiralConv\cite{Chen_Liu_Dong_Zhang_Ma_Xiong_Zhang_Guo_2022} & 6.8& 6.9& 0.724& 0.977\\
\hline
Dynamic Spiral Convolution ($K$=8) &\textbf{6.6}&\textbf{6.7}& \textbf{0.734}& \textbf{0.978} \\
\hline
\end{tabular}
\end{center}
\caption{Ablation study of 3D decoding. }
\label{tab2}
\end{table*}

\begin{table}[t]
\begin{center}
\begin{tabular}{|c|c|c|c|c|}
\hline
$K$ & PJ$\downarrow$ & PV$\downarrow$ & F@5$\uparrow$ & F@15 $\uparrow$  \\
\cline{1-5}
2& 6.7& 6.9& 0.728&0.976 \\
\hline
4 &6.7& 6.8& 0.730&0.976 \\
\hline
8 &\textbf{6.6}&\textbf{6.7}& \textbf{0.734}& \textbf{0.978} \\
\hline
16 &6.7& 6.8& 0.731&0.977 \\
\hline
\end{tabular}
\end{center}
\caption{Ablation study of Dynamic Spiral Convolution. }
\label{tab3}
\end{table}

\begin{table*}[t]

\begin{center}
\begin{tabular}{|c|c|c|c|c|c|c|c|c|}
\hline
& Methods &Venue & Backbone & PJ$\downarrow$ & PV$\downarrow$ & F@5$\uparrow$ & F@15 $\uparrow$ & FPS$\uparrow$ \\
\cline{1-9}
\multirow{5}{*}{\begin{tabular}[c]{@{}c@{}}Non Real-Time  \\ Methods\end{tabular}  } 
&METRO \cite{lin2021end} &CVPR 21 & HRNet &6.7& 6.8& 0.717&0.981&- \\
\cline{2-9}
&MeshGraphormer \cite{lin2021-mesh-graphormer} &ICCV 21 & HRNet &5.9& 6.0& 0.764&0.986&- \\
\cline{2-9}
&PointHMR \cite{PointHMR}&CVPR 23 & HRNet &6.1& 6.6& 0.720&0.984&- \\
\cline{2-9}
&HaMeR \cite{pavlakos2024reconstructing}&CVPR 24 & ViTPose &6.0& 5.7& 0.785&0.990&- \\
\cline{2-9}
&HaMeR-170k \cite{pavlakos2024reconstructing}&CVPR 24 & ViTPose &6.1& 5.8& 0.782&0.990&- \\
\cline{1-9}

\multirow{19}{*}{\begin{tabular}[c]{@{}c@{}} Real-Time  \\ Methods\end{tabular}  }& FreiHAND \cite{zimmermann2019freihand} &ICCV 19 & ResNet50 & 11.0& 10.9& 0.516&0.934& -\\
\cline{2-9}
&ObMan \cite{hasson19_obman} &CVPR 19 & ResNet18 & -& 13.2& 0.436&0.908&20 \\
\cline{2-9}
&Boukhayma et al. \cite{boukhayma20193d} &CVPR 19 & ResNet50 & -& 13.0& 0.435&0.898&- \\
\cline{2-9}
&MobileHand \cite{MobileHand:2020} & ICONIP 20 & MobileNetV3 & - & 13.1& 0.439&0.902&- \\
\cline{2-9}
&YoutubeHand \cite{kulon2020weakly} &CVPR 20 & ResNet50 &8.4& 8.6& 0.614&0.966&- \\
\cline{2-9}
&I2L-MeshNet \cite{moon2020i2l} &ECCV 20 & ResNet50 &7.4&7.6& 0.681& 0.973&33 \\
\cline{2-9}
&Pose2Mesh \cite{Choi_2020_ECCV_Pose2Mesh} &ECCV 20 & HRNet &7.7&7.8& 0.674& 0.969&22 \\
\cline{2-9}
&HIU-DMTL \cite{zhang2021hand} &ICCV 21 & Hourglass &7.1& 7.3& 0.699&0.974&- \\
\cline{2-9}
&CMR \cite{Chen_Liu_Ma_Chang_Wang_Chen_Guo_Wan_Zheng_2021} &CVPR 21 & ResNet50 &6.9& 7.0& 0.715&0.977&30 \\
\cline{2-9}
&I2UV-HandNet \cite{chen2021i2uv} &ICCV 21 & ResNet50 &6.7& 6.9& 0.707&0.977&- \\
\cline{2-9}
&Tang et al. \cite{tang2021towards}&ICCV 21 & ResNet50 &6.7& 6.7& 0.724&\textbf{0.981}&39 \\
\cline{2-9}
&MobRecon \cite{chen2022mobrecon}&CVPR 22 & GhostStack &8.8& 9.1& 0.597&0.960&- \\
\cline{2-9}
&MobRecon \cite{chen2022mobrecon}&CVPR 22 & DenseStack &6.9& 7.2& 0.694&0.979&- \\
\cline{2-9}
&MobRecon \cite{chen2022mobrecon}&CVPR 22 & ResNet18 &6.7& 6.8& 0.727&0.979&69 \\
\cline{2-9}
&MobRecon \cite{chen2022mobrecon}&CVPR 22 & ResNet34 &6.6& 6.7& 0.730&0.979&43 \\
\cline{2-9}
&CLIP-Hand3D \cite{guo_clip_hand}&ACM MM 23 & ResNet50 &6.6& 6.7& 0.728&\textbf{0.981}&\textbf{77} \\
\cline{2-9}
&FastViT \cite{vasufastvit2023} &ICCV 23 & FastViT &6.6& 6.7& 0.722&\textbf{0.981}&- \\
\cline{2-9}

&\textbf{M3DHMR}&\textbf{Ours} & \textbf{ResNet18}  &\textbf{6.6}&\textbf{6.7}&\textbf{0.734} & 0.978&42 \\
\cline{2-9}
&\textbf{M3DHMR}&\textbf{Ours} & \textbf{ResNet34}  &\textbf{6.5}&\textbf{6.6}&\textbf{0.739} & 0.980&28 \\
\hline

\end{tabular}

\end{center}
\caption{Results on the FreiHAND dataset. }
\label{tab4}
\end{table*}

\subsection{Ablation Study}
\paragraph{DSC Layer and ROI Layer}
Tab. \ref{tab1} illustrates the influence of the two layers on the network performance. Compared to the DSC Layer, ROI Layer employs dilated spiral convolution with a larger convolution kernel, thereby enabling a larger receptive field. Furthermore, by limiting the receptive field to the hand region where the convolution centre is located, ROI Layer mitigates the impact of mesh vertices with low correlation across regions. As can be seen from the first and second rows of Tab. \ref{tab1}, the ROI Layer results in a decrease in PV by 0.1 mm, an increase in F@5 by 0.003, and an increase in F@15 by 0.01. The DSC Layer employs a mechanism of adjusting the weights, which allows the convolution to adapt to hand features in different positions. As shown in the first and third rows of Tab. \ref{tab1}, the performance of the dynamic spiral convolution is superior to that of the normal spiral convolution, resulting in a decrease in PJ and PV by 0.1mm and 0.2mm, respectively, and an increase in F@5 and F@15 by 0.008 and 0.011, respectively. The combination of both layers yields the best reconstruction results, as can be seen in the last row of the table.

\paragraph{Methods of 3D Decoding}
As illustrated in Tab. \ref{tab2}, Dynamic Spiral Convolution decreases PJ and PV of the 3D decoder and increases F@5 compared with SpiralConv++ and Depth-Separable SpiralConv. The results demonstrate that adaptively adjusting the convolution weights for vertices at different positions significantly enhances the reconstruction outcomes.

\paragraph{Kernel Size of Dynamic Spiral Convolution}
Tab. \ref{tab3} demonstrates the effect of DSC Layers at different values of $K$. It can be observed that the model exhibits optimal performance with $K$=8. A value that is too small may result in limited expression and insufficient adaptive ability. Conversely, a value that is too large may lead to an excessive focus on detailed features, which may result in the neglect of some main features and an adverse effect on the model's generalization ability.

\subsection{Visualization results of our M3DHMR}
As shown in Fig. \ref{fig1}, the qualitative evaluation results on the wild data demonstrate that our approach achieves high reconstruction accuracy in the situation of complex pose (first and second rows), self-occlusion (third and fourth rows) and object occlusion (fifth and sixth rows).

We randomly select some samples from the test set of FreiHAND and show the comprehensive qualitative results of our predicted 2D pose, front view mesh and side view mesh in Fig. \ref{fig4}. Overcoming challenges including complex poses, object occlusion, truncation and poor lighting, our method generates accurate 2D poses and 3D meshes. More results are shown in Fig. S1 in the Supplemental Material.

\begin{figure*}[t] %!t
\centering
\includegraphics[scale=0.72]{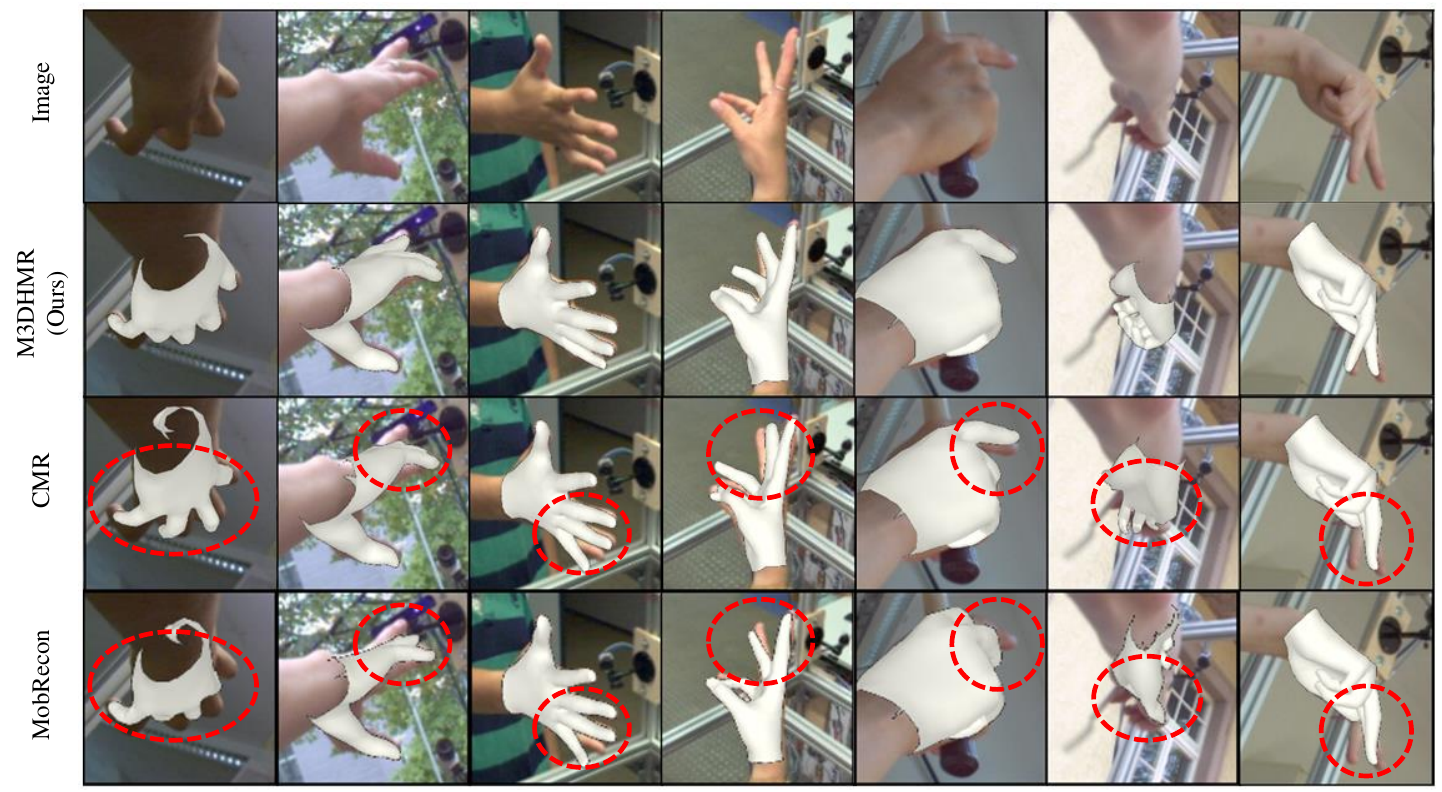}
\caption{Qualitative comparisons of M3DHMR, CMR and MobRecon for predicting front view mesh on the test set of FreiHAND. The shortcomings of the latter two methods are marked with red circles.}
\label{fig5}
% \vspace{-1.0em}
\end{figure*}

\subsection{Comparisons with Contemporary Methods}
In Tab. \ref{tab4}, we compare our method with several state-of-the-art approaches on hand mesh recovery task using the FreiHAND dataset. Since the implementation of Stacked-ResNet in MobRecon is not available, we implement Stack-ResNet18 and fully reproduce the original results of MobRecon. Considering that Stack-ResNet34 can be achieved by adjusting hyperparameters of Stack-ResNet18, we select Stack-ResNet18 and Stack-ResNet34 as backbones for fair comparison between MobRecon and M3DHMR. Our ResNet34-based M3DHMR achieves the best performance among real-time methods on camera-space mesh recovery. Specifically, M3DHMR reaches lowest 6.5mm PA-MPJPE, lowest 6.6mm PA-MPVPE and highest F@5 of 0.739. Additionally, its F@15 of 0.980 also outperforms most compared approaches, only slightly lower than CLIP-Hand3D and FastViT. Furthermore, a qualitative comparison of our M3DHMR with two other state-of-the-art real-time methods is presented in Fig. \ref{fig5}. It can be observed that our method demonstrates superior performance in terms of detail reconstruction. More results are shown in Fig. S2 in the Supplemental Material.

\section{Conclusion and Discussion}
In this paper, we provide a monocular 3D hand mesh recovery method named M3DHMR. Firstly, M3DHMR adopts a stacked hourglass structure to introduce 2D cues. Then, a spiral decoder with DSC Layers and a ROI Layer lifts the 2D cues to 3D mesh. By adaptively extracting local features at different positions and fully exploring the global features in each hand region, our model achieves the state-of-the-art performance on real-time 3D hand mesh recovery tasks when evaluated on the FreiHAND dataset. 

However, there is still some room for improvement in depth ambiguity. In the future, we intend to explore the potential of incorporating data distribution priors to mitigate depth ambiguity by introducing generative models, such as VAE and diffusion models. 

\section*{Ethics}

This study utilizes the publicly available FreiHAND dataset for monocular 3D hand mesh recovery. The original dataset collection was conducted in accordance with ethical research guidelines, with informed consent obtained from all participants. No additional data collection was performed in this work. We affirm our commitment to responsible data usage practices and emphasize the importance of maintaining ethical vigilance in all applications of this research.

% \newpage
{\small
\bibliographystyle{ieee}
\bibliography{references}
}

\end{document}